\documentclass[conference]{IEEEtran}
\IEEEoverridecommandlockouts

\usepackage{eso-pic}

\AddToShipoutPictureFG*{%
	\AtPageUpperLeft{%
		\raisebox{-0.45in}{%
			\makebox[\paperwidth]{%
				\parbox{0.95\paperwidth}{\fontsize{7}{8}\selectfont
					Preprint version.
				}%
			}%
		}%
	}%
}

\usepackage{eso-pic}
\AddToShipoutPictureFG*{%
	\AtPageLowerLeft{%
		\raisebox{0.28in}{%
			\makebox[\paperwidth]{%
				\parbox{0.95\paperwidth}{\centering\fontsize{7}{8}\selectfont
					Preprint Version. This work has been submitted to the IEEE for possible publication. Copyright may be transferred without notice, after which this version may no longer be accessible.
				}%
			}%
		}%
	}%
}

\usepackage[T1]{fontenc}
\usepackage[utf8]{inputenc}
\usepackage{graphicx}
\usepackage{amsmath}
\usepackage{booktabs}
\usepackage{multirow}
\usepackage{url}
\usepackage{xcolor}

\usepackage{cite}
\usepackage{amsmath,amssymb,amsfonts}
\usepackage{algorithmic}
\usepackage{textcomp}

\title{LLM Agents Perform Controlled Experiments Using Simulation Models}

\author{%
\IEEEauthorblockN{Yuchen Xia, Michael Weyrich, Nasser Jazdi, Johannes St\"umpfle, Johannes Sigel, Akshay Narla}
\IEEEauthorblockA{Institute for Industrial Automation and Software Engineering\\University of Stuttgart, Stuttgart, Germany\\
\{yuchen.xia $|$ michael.weyrich $|$ nasser.jazdi $|$ johannes.stuempfle $|$ johannes.sigel $|$ akshay.narla\}@ias.uni-stuttgart.de}
\and
\IEEEauthorblockN{Gavin K. Reynolds\IEEEauthorrefmark{1}, Anna Jawor-Baczynska\IEEEauthorrefmark{2}, Pol Llopart\IEEEauthorrefmark{3}}
\IEEEauthorblockA{AstraZeneca\\
\IEEEauthorrefmark{1}Sustainable Innovation \& Transformational Excellence (xSITE), Pharmaceutical Technology \& Development, Operations\\
\IEEEauthorrefmark{2}Chemical Development, Pharmaceutical Technology \& Development, Operations\\
\IEEEauthorrefmark{3}Data Analytics \& AI (DA\&AI), Operations IT\\
\IEEEauthorrefmark{1,2}Macclesfield, UK; \IEEEauthorrefmark{3}Barcelona, Spain\\
\{Gavin.Reynolds $|$ Anna.Jawor-Baczynska $|$ Pol.Llopart\}@astrazeneca.com}
}

\begin{document}
\maketitle

\begin{abstract}
Large language models (LLMs) have shown strong capabilities in reasoning, planning, and tool use, but many scientific and engineering tasks require more than plausible text and code generation. They require understanding how a system responds to intervention, which in practice depends on controlled experimentation. In this work, we propose a multi-agent framework that enables LLM agents to conduct controlled experiments with scientific simulation models for pharmaceutical process design. Given a user query and a baseline configuration, the system constructs a structured task representation, designs experiments, executes comparative simulation, interprets the resulting outcomes, and synthesizes evidence-based recommendations for process parameter optimization. By coupling language models with high-fidelity simulation models in an interactive agent framework, the proposed system supports reasoning through intervention, comparison, and observation. As a result, it produces more specific and actionable outputs than language-only reasoning. In an industrial application setting, this advantage is reflected in higher output specificity as well as improved user-rated correctness and helpfulness. Ablation studies and visualized case analyses further demonstrate the effectiveness and practical utility of simulation-integrated experimental reasoning.
\end{abstract}


\section{Introduction}

Large language models (LLMs) have shown promising capabilities in multi-step reasoning, planning, and tool use. However, many scientific and engineering tasks require more than plausible text generation. They require determining how a system responds to intervention, which in practice means reasoning through controlled comparison rather than through language generation alone.

Controlled experimentation is a central mechanism of scientific inquiry and engineering problem-solving. To understand how a process should be improved, one typically formulates a hypothesis, varies a selected factor while keeping other conditions fixed, observes the resulting change, and compares it against a reference condition. This logic is essential for identifying causal effects and for producing conclusions that are specific, testable, and actionable.

Current LLM-based systems do not naturally operate in this mode. Even when they are equipped with external tools, they are often used to retrieve information or execute isolated functions, rather than to carry out structured experimental comparison. As a result, their outputs may remain suggestive rather than evidential, especially in tasks where reliable conclusions depend on comparing outcomes under controlled intervention.

This issue is particularly relevant in scientific and industrial applications, where important knowledge is often embodied in simulation models that capture system dynamics under varying operating conditions. In pharmaceutical process design, for example, simulation models can provide a practical basis for evaluating candidate process modifications and their consequences.

Motivated by this setting, we investigate how LLMs can be placed in a simulation-based experimental environment for scientific reasoning. Rather than treating simulation as a passive auxiliary tool, we consider it as an environment in which hypotheses can be tested through controlled intervention and comparison. This perspective provides the basis for the framework developed in this work.
\begin{figure*}[t]
	\centering
	\includegraphics[width=\textwidth]{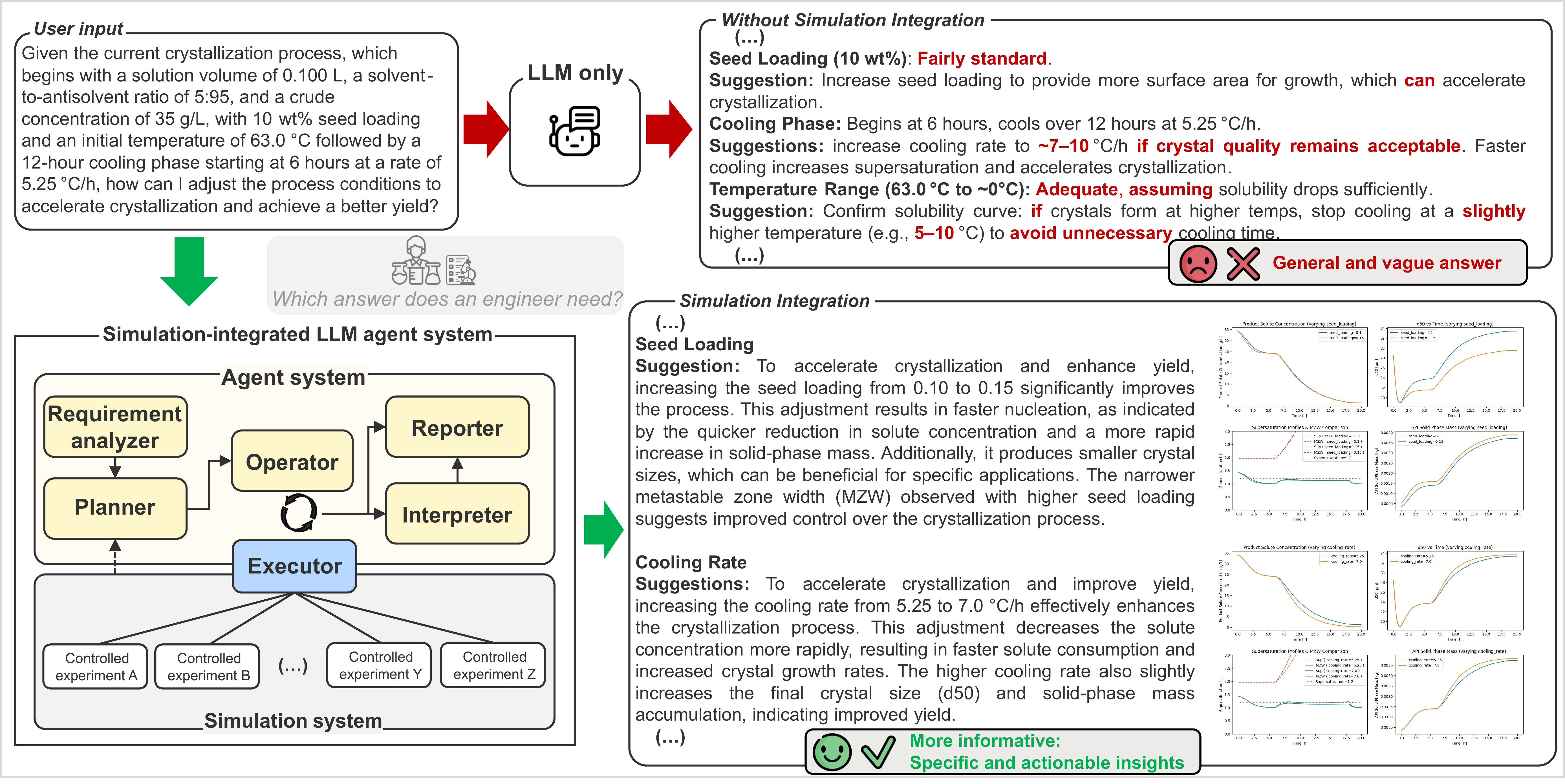}
		\vspace{-2.2em}
	\caption{Graphical abstract. Simulation-integrated agent system delivers precise, evidence-based insights to support engineering decision-making.}
	\label{fig:graphical-abstract}
\end{figure*}

\section{Related Work}

\subsection{Related application fields and use cases}

Across scientific and engineering domains, LLMs have increasingly been applied to domain-specific workflows. In chemistry, prior work has demonstrated tool-augmented chemical reasoning in ChemCrow \cite{Bran2024}, autonomous experiment design and execution in Coscientist \cite{Boiko2023}, and end-to-end synthesis development in LLM-RDF \cite{Ruan2024}. In simulation-centered engineering settings, recent systems have supported OpenFOAM-based computational fluid dynamics setup and refinement in OpenFOAMGPT \cite{Pandey2025}, natural-language-to-CFD automation in AutoCFD \cite{Dong2025}, and broader end-to-end simulation research workflows in ASA \cite{Liu2025}.

These studies demonstrate the growing applicability of LLMs to domain-specific scientific and engineering tasks. However, most existing systems emphasize retrieval, workflow automation, code generation, or simulation setup, rather than using simulation models as an environment for controlled experiments and comparative reasoning. As a result, simulation is typically treated as a tool for task execution, not as an experimental substrate for testing hypotheses through variable intervention and outcome interpretation.

\subsection{Reasoning mechanism and agent framework}

Prior studies have shown that LLMs can perform iterative reasoning and interact with external tools, as in ReAct \cite{Yao2023ReAct}, Toolformer \cite{Schick2023}, and Gorilla \cite{Patil2024}. Multi-agent frameworks such as AutoGen \cite{Wu2024}, HuggingGPT \cite{Shen2023}, and CAMEL \cite{Li2023CAMEL} assign functional roles to agents for solving general and domain-specific tasks. Communicative Agents \cite{Qian2024,Xia2023} further show how multi-round collaboration can improve performance in structured workflows.

More recent work has emphasized explicit role decomposition and the separation of planning from execution\cite{Xia2026Integrating}, as in ConAgents \cite{Shi2024} and Plan-And-Act \cite{Erdogan2025}. Reliable tool use has also been improved through clearer and more standardized tool descriptions, as in EASYTOOL \cite{Yuan2025}. These works provide important design principles for building structured LLM agent systems. However, they do not directly address how such agent architectures can be organized around controlled experimental reasoning over scientific simulation models.

\subsection{Simulator and tool integration}

A growing body of work has explored the integration of LLMs with tools, software environments, and simulation-related components. Existing systems have shown that LLM agents can access tools \cite{Zhao2024} and Web APIs \cite{Qin2024}, and can be connected to simulation-oriented workflows such as OpenFOAM-based CFD environments \cite{Pandey2025} and broader automated simulation research pipelines \cite{Liu2025}. Multi-agent systems have also been used to formulate, execute, and validate physics-based simulations, for example in mechanics problems in MechAgents \cite{Ni2024} and in protein design and analysis in ProtAgents \cite{Ghafarollahi2024}.

At the same time, many simulator-related LLM studies remain situated in simplified, embodied, or sandbox-style environments, including 3D simulators \cite{Huang2022}, rule-based game settings \cite{Akata2025}, and virtual environments used for behavioral exploration \cite{Wang2024Voyager,Xia2024}. In such settings, simulation often serves as a testbed for action generation or planning behavior rather than as a high-fidelity scientific environment for controlled comparison and evidence-grounded reasoning.

\begin{figure*}[t]
	\centering
	\includegraphics[width=\linewidth]{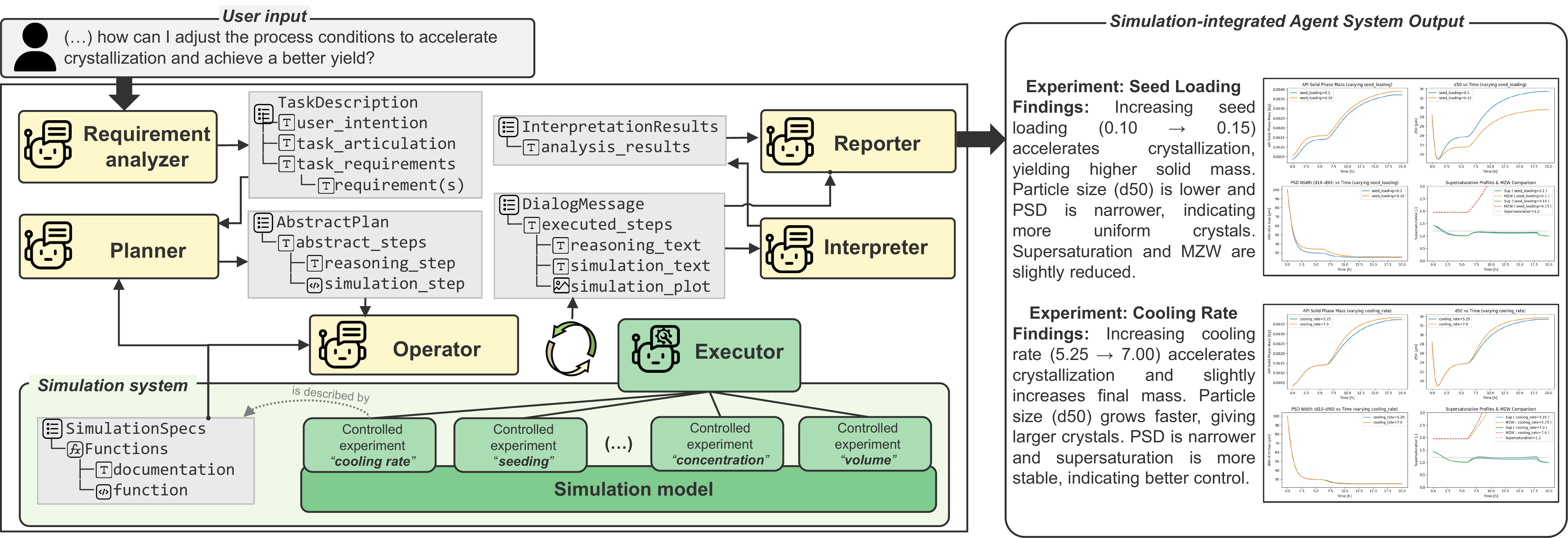}
	\vspace{-2.2em}
	\caption{Information flow in the proposed simulation-integrated agent framework.}
	\label{fig:framework}
\end{figure*}

\subsection{Contributions beyond prior work}

The main contributions of this work are as follows:

\textbf{Scientific reasoning as structure.} The proposed agent architecture organizes reasoning according to a scientific problem-solving paradigm. The system decomposes a task into subgoals and requirements, designs controlled experiments to test hypotheses, observes the resulting outcomes, and synthesizes these observations into conclusions. This structure supports systematic and verifiable reasoning for engineering applications.

\textbf{Scientific simulation.} While prior work has demonstrated the feasibility of grounding LLMs with simulation software, often in simplified or game-like environments, our framework is designed to interact with high-fidelity scientific simulations that reproduce dynamic physical and chemical phenomena. This enables the system to draw on complementary knowledge from both the simulation model and the language model.

\textbf{Application-driven design.} The framework is developed under real-world application constraints. By formalizing the reasoning process as a directed graph, the system provides clear observability and visualization of intermediate reasoning artifacts, which supports practical industrial use.

\section{Method}

Fig.~\ref{fig:framework} illustrates the proposed simulation-integrated agent framework. The system processes a user query through a pipeline of specialized agents and generates recommendations for engineer users. 

The overall design is organized as a structured information-processing workflow in which intermediate reasoning artifacts are explicitly generated, transformed, and passed between agents. In the context of this work, this workflow enables the LLM-based system to carry out controlled experimental reasoning: it analyzes the task, identifies relevant intervention variables, plans simulation-based comparisons, executes parameterized experiments, interprets the outcomes, and synthesizes the results into a final recommendation.

\subsection{Agent Framework Design}

The proposed framework consists of six distinct agents. Five of them are LLM-driven agents, each guided by a dedicated prompt, while one, the Executor Agent, is implemented as a rule-based software component. Each agent is responsible for a specific functional role within the overall pipeline.

The Requirement Analyzer Agent receives the user input, interprets the underlying user intention, and reasons over it to produce a task articulation together with a structured list of requirements. These outputs jointly form the task description, which serves as the basis for downstream reasoning.

The Planner Agent takes the task description as input and performs step-by-step reasoning to generate an abstract plan without committing to concrete execution details. The Planner is provided with a list of available simulation functions, referred to as Simulation Specs, where each function is annotated with a concise description. Based on the task requirements and available simulation capabilities, the Planner produces an abstract plan that outlines the logical sequence of reasoning steps and identifies where simulation functions are relevant. The Planner is explicitly instructed to remain at the abstract level and not to generate detailed execution logic or outcomes.

The Interactive Operator Agent receives the abstract plan and concretizes it through detailed reasoning. In particular, it operationalizes those plan steps that require simulation-based comparison. Whenever the Operator reaches a step involving a simulation function, it generates Python code with fully specified input parameters and delegates the execution to the Executor. After receiving the execution results, it resumes subsequent reasoning. In this way, the Operator converts abstract reasoning steps into executable controlled simulation experiments.

The Interactive Executor Agent executes modularized simulation functions in a controlled Python interpreter environment. It returns both textual and graphical outputs and interacts directly with the Operator by providing execution feedback. As a deterministic software component, it is responsible for reliable tool execution rather than language reasoning.

The Interpreter Agent semantically interprets simulation-generated plots using a vision-capable LLM. It produces textual insights from visual simulation outputs and summarizes the outcomes of the parameterized experiments conducted through the simulation model.

The Reporter Agent observes the overall reasoning and execution process, aggregates the textual outputs generated by the previous agents, and produces a user-facing response that summarizes both the task-solving process and the final results.

Taken together, these agents organize the system into a structured reasoning pipeline that progressively transforms a user query into explicit intermediate artifacts, simulation-based evidence, and final recommendations.

\subsection{Interactive Interface Between LLM Agents and the Simulation Model}

\begin{figure}[t]
	\centering
	\includegraphics[width=\linewidth]{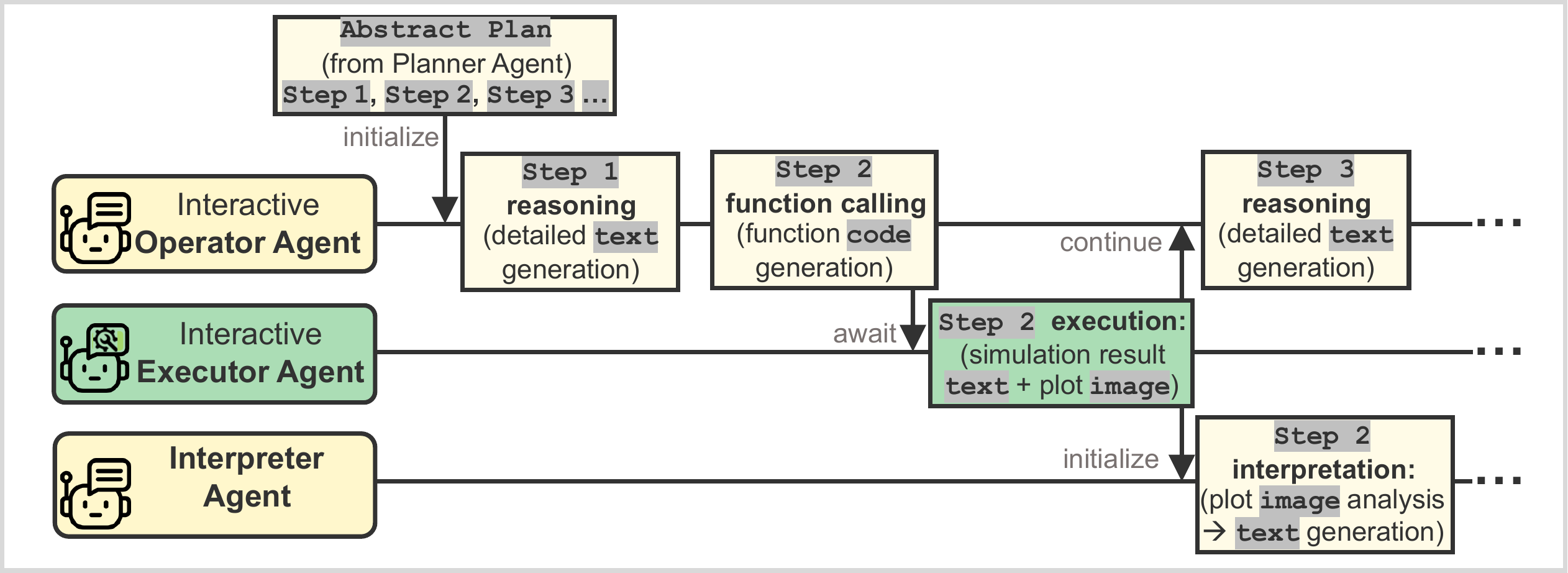}
	\caption{Operator--Executor--Interpreter interaction protocol during simulation invocation.}
	\label{fig:interaction-protocol}
\end{figure}

The interface between the agent system and the simulation model is realized through the coordinated interaction of three agents: the Operator, the Executor, and the Interpreter, as shown in Fig.~\ref{fig:interaction-protocol}. The Operator and Executor form a two-agent dialogue loop for simulation invocation, while the Interpreter analyzes the visual outputs produced by the simulation.

Starting from the abstract plan, the Operator expands each relevant plan step into concrete reasoning text. When a simulation experiment is required, it generates the corresponding function-call code and sends it to the Executor. The Executor runs the code, obtains both textual and visual outputs from the simulation functions, and returns the textual execution results to the Operator, which then continues its reasoning. In parallel, the visual outputs, such as plots, are passed to the Interpreter, which extracts semantic insights in textual form.

This interaction protocol enables an adaptive and fault-tolerant tool-use process. If a function call fails, the deterministic Executor returns an error message. The Operator can then revise the function call and retry generation until the execution succeeds or a predefined retry limit is reached. This design allows the system to recover from malformed or incomplete tool invocations while maintaining the continuity of the reasoning process.

\subsection{Structured Graph-Based Visualization}

\begin{figure*}[t]
	\centering
	\includegraphics[width=\linewidth]{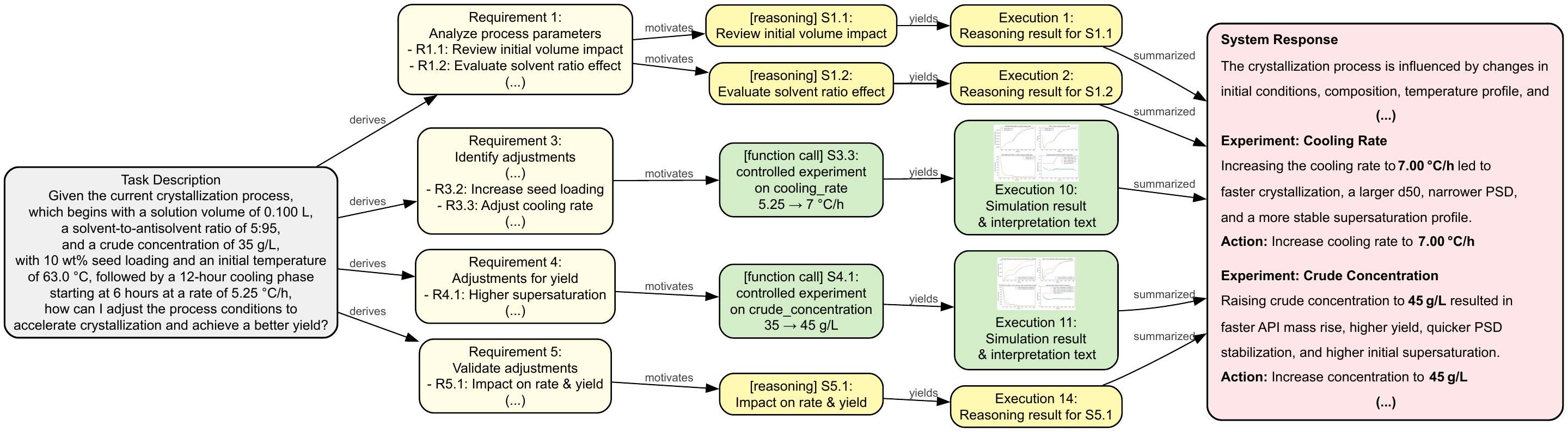}
	\caption{Graph illustrating the reasoning trajectories from user task to system response.}
	\label{fig:reasoning-graph}
\end{figure*}

To systematically observe and analyze the generated content from the proposed system, we visualize the reasoning trajectories, as shown in Fig.~\ref{fig:reasoning-graph}.
The operation of the proposed system can be represented as a directed graph that captures the flow of reasoning from the initial task description to the execution and interpretation of simulation results. Nodes denote discrete generated artifacts, such as task descriptions, requirements, planning steps, simulation calls, and outputs, while edges capture their logical or procedural dependencies. The graph uses four types of edges to represent the reasoning flow and the relationships between intermediate system artifacts: \emph{derives}, \emph{motivates}, \emph{yields}, and \emph{summarized}. Together, these nodes and edges form the system’s reasoning trajectories, providing clear observability and diagnosability.

\section{Experiments}

This section evaluates the proposed system from both quantitative and qualitative perspectives. We conduct a comparative analysis to understand how simulation integration, requirement analysis, and the agent-based framework each affect the reasoning process, output specificity, and practical usability.

\subsection{Experimented System Variants}

We evaluate four system configurations, with all agent-based variants powered by GPT-4o:

\begin{itemize}
	\item \textbf{Full system (agent + simulation integration + requirement analysis):} complete pipeline
	\item \textbf{No simulation:} same system, but with simulation functions disabled
	\item \textbf{No requirements:} same system, but with the Requirement Analyzer disabled
	\item \textbf{LLM only:} a vanilla LLM (GPT-4o) directly prompted with the user task
\end{itemize}

These variants allow us to isolate the contribution of integrated simulation, requirement structuring, and the multi-agent workflow.

\subsection{Tasks}

All system variants are evaluated on a test set consisting of five task scenarios in pharmaceutical crystallization process design (see Appendix: Reasoning Trajectories). Each task specifies a user-defined baseline experiment configuration together with a task goal.

Given a natural-language user query $q$ with an optimization goal $g$ (e.g., maximizing yield), and a baseline process configuration represented by a parameter vector $\mathbf{x}=(x^{(1)},x^{(2)},\ldots,x^{(n)})$, the system determines how to improve the outcome by conducting controlled simulation experiments.

The simulation-integrated agent system is expected to perform the following sequence of operations:

\begin{itemize}
	\item Identify the elements of the baseline parameter vector $\mathbf{x}$ that are relevant to the optimization goal $g$
	\item For each selected element $x^{(i)}$ in $\mathbf{x}$, generate a controlled perturbation 
	\begin{equation}
		x^{(i)'} = x^{(i)} + \delta^{(i)}
	\end{equation}
	while keeping all other elements unchanged, yielding
	\begin{equation}
		\mathbf{x}' = (x^{(1)},\ldots,x^{(i)'},\ldots,x^{(n)})
	\end{equation}
	\item Using the simulation function $f(\mathbf{x})$, compare the baseline result $f(\mathbf{x})$ with the perturbed result $f(\mathbf{x}')$
	\item Summarize the findings from these controlled comparisons and recommend an improved parameter vector $\mathbf{x}^{*}$ such that $f(\mathbf{x}^{*})$ is closer to $g$
\end{itemize}

This pipeline reflects the principle of a structured scientific inquiry: hypothesize $\rightarrow$ intervene $\rightarrow$ observe $\rightarrow$ analyze $\rightarrow$ report. It also serves as a fundamental unit for solving more complex optimization problems that require clear and specific conclusions.

\subsection{Simulation Model}

The simulation model used in this work is a proprietary industrial asset. It is built on first-principles equations combined with semi-empirical formulations calibrated through fitted parameters. The model consists of differential and algebraic equations and is capable of reproducing key system dynamics, including mass transfer, phase transitions, and time-evolving material distributions in a crystallization process.

The model is integrated into the agent system through software interfaces, where simulation capabilities are exposed as callable functions with defined specifications accessible to the LLM agents, as illustrated in Fig.~\ref{fig:framework}.

\subsection{Data Collection}

For each combination of system variant and task, we collect the following data:

\begin{itemize}
	\item Final outputs, together with intermediate reasoning steps and simulation plots. These are further represented as reasoning trajectory graphs composed of nodes and edges (see Fig.~\ref{fig:reasoning-graph}), with detailed examples provided in Appendix: Reasoning Trajectories
	\item Evaluation data, including (i) technical metrics that quantify linguistic uncertainty, such as lexical statistics and fuzziness scores, and (ii) user annotations assessing the correctness and practical usefulness of the system outputs
\end{itemize}

\section{Results Evaluation}

\subsection{Qualitative Result Comparison}

Simulation enables precise, evidence-grounded recommendations. To illustrate the effect of simulation integration on reasoning quality, we compare outputs from system variants with and without simulation integration on a representative task, as shown in the following example:
\vspace{-1.5em}
\begin{center}
	\fbox{
		\begin{minipage}{0.95\linewidth}
			\textbf{User input:} ``Under the current conditions, I want to grow crystals with a median size of 50 $\mu$m. What process modifications would achieve this goal?''
		\end{minipage}
	}
\end{center}

When given this user input, a vanilla LLM (powered by GPT-4o without simulation integration) produces a vague and heuristic response, indicating that the answer is fuzzy but safe and therefore has high entropy and uncertainty:

\begin{center}
	\fbox{%
		\begin{minipage}{\linewidth}
			\small
			\textbf{Without simulation:} ``The initial temperature of 63.0 $^\circ$C and a cooling rate of 5.25 $^\circ$C/h \textcolor{red}{may} lead to rapid supersaturation and smaller crystals. A slower cooling rate \textcolor{red}{could} allow more time for crystal growth. Implementing a slower cooling rate, such as 3 $^\circ$C/h, \textcolor{red}{may} allow more time for crystal growth and \textcolor{red}{potentially} achieve larger crystals.''
		\end{minipage}%
	}
\end{center}

The proposed system generates the following specific, evidence-based answer grounded in the simulation results:
\vspace{-1.5em}
\begin{center}
	\fbox{%
		\begin{minipage}{\linewidth}
			\small
			\textbf{Full system:} ``A higher cooling rate of 5.25 results in a greater API solid-phase mass, larger median particle sizes (d50), and a narrower particle size distribution (PSD) width compared to a cooling rate of 3.0. The increased cooling rate also leads to higher supersaturation and a larger metastable zone width (MZW), promoting faster nucleation and growth. To achieve a median particle size of 50 $\mu$m, consider further increasing the cooling rate beyond 5.25.''
		\end{minipage}%
	}
\end{center}

This contrast illustrates a shift from vague heuristics to actionable and precise reasoning enabled by simulation integration. A more detailed qualitative illustration of the system reasoning behaviors is provided in the Appendix.

\subsection{Quantitative Evaluation Metrics}
To assess how simulation integration enhances language model reasoning, we adopt three complementary evaluation perspectives. First, we measure reasoning quality using two core metrics: specificity and correctness. Second, we evaluate practical usefulness through user-rated helpfulness scores. Third, we assess the simulation-calling behavior of the system, including the precision and recall of simulation calls made by the LLM agents, as well as whether the agents’ hypotheses are validated by simulation outcomes.

\subsubsection{Metrics for Reasoning Specificity}

We use two metrics to evaluate reasoning specificity.

\begin{itemize}
	\item \textbf{Linguistic fuzziness analysis:} the frequency per 1,000 words of expressions signaling uncertainty, including vague modifiers, hedging structures, range expressions, and weak logical connectives. Examples include vague modifiers such as ``somewhat,'' ``likely,'' and ``may''; hedging structures such as ``if,'' ``would,'' and ``could''; range expressions such as ``5 to 10 $^\circ$C''; and weak logical connectives such as ``can,'' ``may,'' and ``might''.
	\item \textbf{LUCI score [0--1]:} a normalized metric \cite{Vincze2014} for quantifying linguistic uncertainty. For example, a score of 0.13 indicates that 13\% of sentences in a paragraph are marked as uncertain, using the implementation from \cite{Meyers2017LUCI}.
\end{itemize}

\subsubsection{Metrics for Reasoning Correctness and Usefulness}

For correctness and usefulness, we do not use automated evaluation methods such as LLM-as-a-Judge, since LLMs lack real-world experience with the specific scenario knowledge and detailed facts required in this application domain. Standard LLM benchmarks are also too general to be applicable in this context.

We therefore adopt user evaluation. Two senior domain specialists review the full reasoning process, provide commentary, and rate the system outputs based on the following two questions:

\begin{itemize}
	\item \textbf{Correctness [1--5]:} How consistent are the results and intermediate reasoning with your empirical knowledge?
	\item \textbf{Helpfulness / Usefulness [1--5]:} How helpful or useful would the system's output be to an end user in a practical production environment?
\end{itemize}

The reported correctness and helpfulness scores are averaged over five distinct task scenarios.

\subsubsection{Metrics for LLM-Agent Invoked Simulation Calls}

In the proposed framework, the Planner Agent is responsible for planning simulation steps, while the Operator Agent parameterizes changes to specific input variables and invokes the corresponding simulation functions. This process is evaluated using two types of metrics.

\begin{itemize}
	\item \textbf{Simulation call precision / recall (Sim. P/R):} Precision is defined as the proportion of simulation calls made by the system that are appropriate. Recall is defined as the proportion of necessary simulation calls that were actually made by the system, where necessity is determined through user annotation.
	\item \textbf{Simulation confirmation (Sim. Conf.):} Simulation confirmation measures the proportion of system-generated hypotheses validated by simulation outputs. A hypothesis is considered confirmed if simulation results are consistent with predictions and move the system closer to the optimization goal. For example, if increasing the cooling rate is hypothesized to accelerate crystallization and simulation results show a corresponding increase, the hypothesis is considered confirmed.
\end{itemize}

\subsection{Evaluation Results with Quantitative Metrics}

Table~\ref{tab:metrics} summarizes reasoning quality metrics across five crystallization tasks, each formulated as an optimization problem. The evaluation includes manual inspection of complete reasoning trajectories represented as graphs, with detailed examples provided in Appendix: Reasoning Trajectories. Specifically, we analyze 17 simulation results generated under the full-system configuration and 30 results from the ablated no-requirement configuration.

\begin{table}[t]
	\centering
	\caption{Reasoning quality metrics under the four system variants (mean $\pm$ standard deviation).}
	\label{tab:metrics}
	\resizebox{\linewidth}{!}{%
		\begin{tabular}{lcccccc}
			\toprule
			System Variant & Fuzzy Words / 1k $\downarrow$ & LUCI $\downarrow$ & Correct $\uparrow$ & Helpful $\uparrow$ & Sim. P/R $\uparrow$ & Sim. Conf. $\uparrow$ \\
			\midrule
			Full System & $13.7 \pm 6.7$ & $0.13 \pm 0.10$ & 4.1 & 4.2 & 94\% / 64\% & 76\% \\
			Without Requirement & $17.7 \pm 5.9$ & $0.17 \pm 0.07$ & 3.4 & 3.8 & 65\% / 69\% & 45\% \\
			Without Simulation & $51.3 \pm 4.7$ & $0.33 \pm 0.09$ & $(< 3.0)$ & $(< 3.0)$ & N.A. & N.A. \\
			LLM-only & $57.1 \pm 14.6$ & $0.36 \pm 0.11$ & $(< 3.0)$ & $(< 3.0)$ & N.A. & N.A. \\
			\bottomrule
	\end{tabular}}
\end{table}

As shown in Table~\ref{tab:metrics}, the Full System outperforms all ablated variants in terms of output specificity, correctness, and helpfulness. Its responses contain the fewest vague expressions, at 13.7 per 1,000 words, and achieve the lowest LUCI uncertainty score, 0.13, indicating more precise and specific reasoning. This is especially important for scientific and engineering tasks. In contrast, the No Simulation and LLM-only variants contain substantially more vague and uncertain expressions, with LUCI scores of 0.33 and 0.36, respectively. While some of these outputs may not directly contradict known facts, their lack of specificity limits their practical usefulness for actionable decision-making.

\subsection{Experiment Hypotheses Confirmed by Simulation (Sim. Conf.)}

The full system achieves 94\% simulation call precision, and 76\% of its reasoning hypotheses are supported by simulation results. For example, the system may generate a hypothesis such as increasing the cooling rate, execute this as a controlled experiment through the simulation model using perturbed parameter inputs, and obtain outputs that confirm measurable improvement toward the goal. This completes the loop of hypothesis $\rightarrow$ experiment execution $\rightarrow$ validation.

Overall, the full system is positively evaluated by domain specialists, achieving an average correctness score of 4.1 and a helpfulness score of 4.2.

\subsection{Ablation on Requirement Analysis (No Requirement)}

Removing the Requirement Analyzer increases simulation recall to 69\%, but reduces simulation precision to 65\%. This suggests that without explicit reasoning over task requirements, the system tends to overuse available simulation functions and produce less targeted simulation calls. The outputs also become less certain according to lexical analysis, and both correctness and usefulness scores decline. This result aligns with prior findings \cite{Yao2023ReAct,Erdogan2025} showing the value of intermediate reasoning steps in complex task solving.

\subsection{Ablation on Simulation Functions (No Simulation)}

The weakest performance is observed when simulation functions are removed. In this setting, all agents rely solely on the LLM's prior knowledge, without access to experimental context. As a result, the outputs become significantly less specific, as reflected by high fuzziness and LUCI scores. Domain users regarded scores below 3.0 (< 3.0) as unusable in this setting because the outputs were too vague to support meaningful insight extraction.

\subsection{Comparison with LLM-Only Prompting}

When given the same query, the LLM-only baseline also produces overly vague responses, as reflected in high fuzziness and LUCI scores. Because the outputs generated without simulation integration contain excessive hedging and fuzzy statements, they are considered unverifiable and unhelpful for engineering users.

\subsection{Summary of Quantitative Evaluation}

These results support the effectiveness of the proposed system design. First, simulation integration enables more precise and evidence-based reasoning. Second, requirement analysis contributes to better reasoning performance. Third, the agent-based architecture successfully combines the strengths of language models and simulation models, organizing the reasoning process into a traceable and testable workflow that mirrors scientific inquiry. As a whole, this design yields outputs that are not only more specific and informative, but also practically useful for engineering optimization tasks.

\section{Discussion and Generalizable Insights}

The overall reasoning quality of the simulation-integrated agent system is governed by a unified principle: approximation fidelity. This principle manifests differently across the two model types. For the simulation model, it refers to physical fidelity, i.e., the degree to which the simulator reproduces real-world dynamics. For the language model, it corresponds to hypothesis correctness, i.e., the extent to which its reasoning proposes correct experimental interventions toward the optimization goal. This unification becomes measurable through whether simulation outcomes confirm the LLM's hypotheses, which can be understood as a form of virtual empirical validation.

The fidelity of the simulation model plays a crucial role. A low-fidelity simulator, such as a game-like environment, may introduce distorted context and misleading details, causing the language model to ground its reasoning in unrealistic or non-existent scenarios and thereby inject noise instead of knowledge. Conversely, a poorly trained LLM may generate incorrect hypotheses and fail to interact meaningfully with the simulation. In both cases, the resulting system lacks the scientific accuracy required for high-stakes engineering applications.

In this application, the developed system is intended as an assistant for decision support. It synthesizes the outputs of both models and presents the agent's reasoning process in a traceable form for the user. The knowledge derived from both models is therefore combined in a complementary way, rather than being dominated by either one. This is important because perfect simulations are rarely achievable, and LLMs do not possess precise training for all specific scenarios. The practical role of the system is therefore not to replace user judgment, but to provide traceable, evidence-grounded support for decision-making under these limitations.

\section{Conclusion}

This work presents a simulation-integrated agent system that enables LLMs to reason through controlled experiments for optimization tasks. Rather than directly generating recommendations, the system analyzes the problem, formulates intervention hypotheses, executes simulation-based comparisons, observes the outcomes, and incorporates the resulting evidence into its final conclusions. In this way, the overall reasoning process follows the logic of scientific inquiry: hypothesize, intervene, observe, and report.

The results show that simulation integration improves reasoning quality by making system outputs more specific, more evidence-grounded, and more useful for practical decision-making. By coupling the hypothesis-generation capability of LLMs with the mechanistic fidelity of scientific simulation, the proposed framework provides a minimal and generalizable approach to optimization-oriented reasoning. Demonstrated here in pharmaceutical process design, the framework suggests a broader direction for future LLM systems: stronger reasoning may depend not only on more capable language models, but also on how they are connected to structured sources of experimental evidence.

\section*{Appendix: Reasoning Trajectories}
\label{app:reasoning-trajectories}

The reasoning trajectories are automatically generated digital artifacts rendered as Scalable Vector Graphics (SGV-images). For clear inspection, please view them in PDF format and zoom in to inspect the details.

\noindent\textbf{Task Sample 1 (Full system)}

\vspace{0.2em}
\noindent\includegraphics[width=\columnwidth]{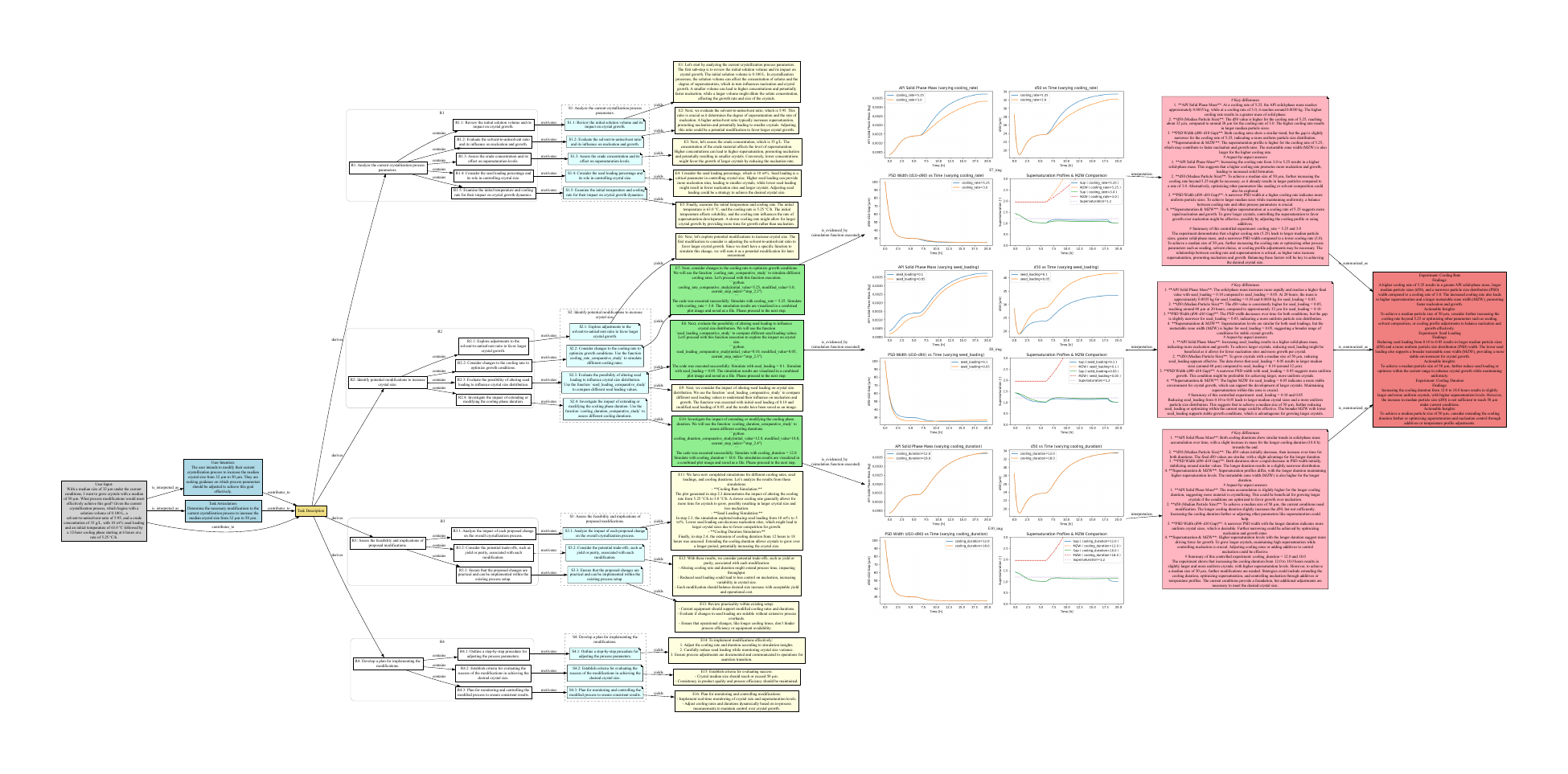}

\noindent\textbf{Task Sample 1 (Ablation: No Requirement Analysis)}

\noindent\makebox[\columnwidth][c]{%
	\includegraphics[width=0.79\columnwidth,angle=-90]{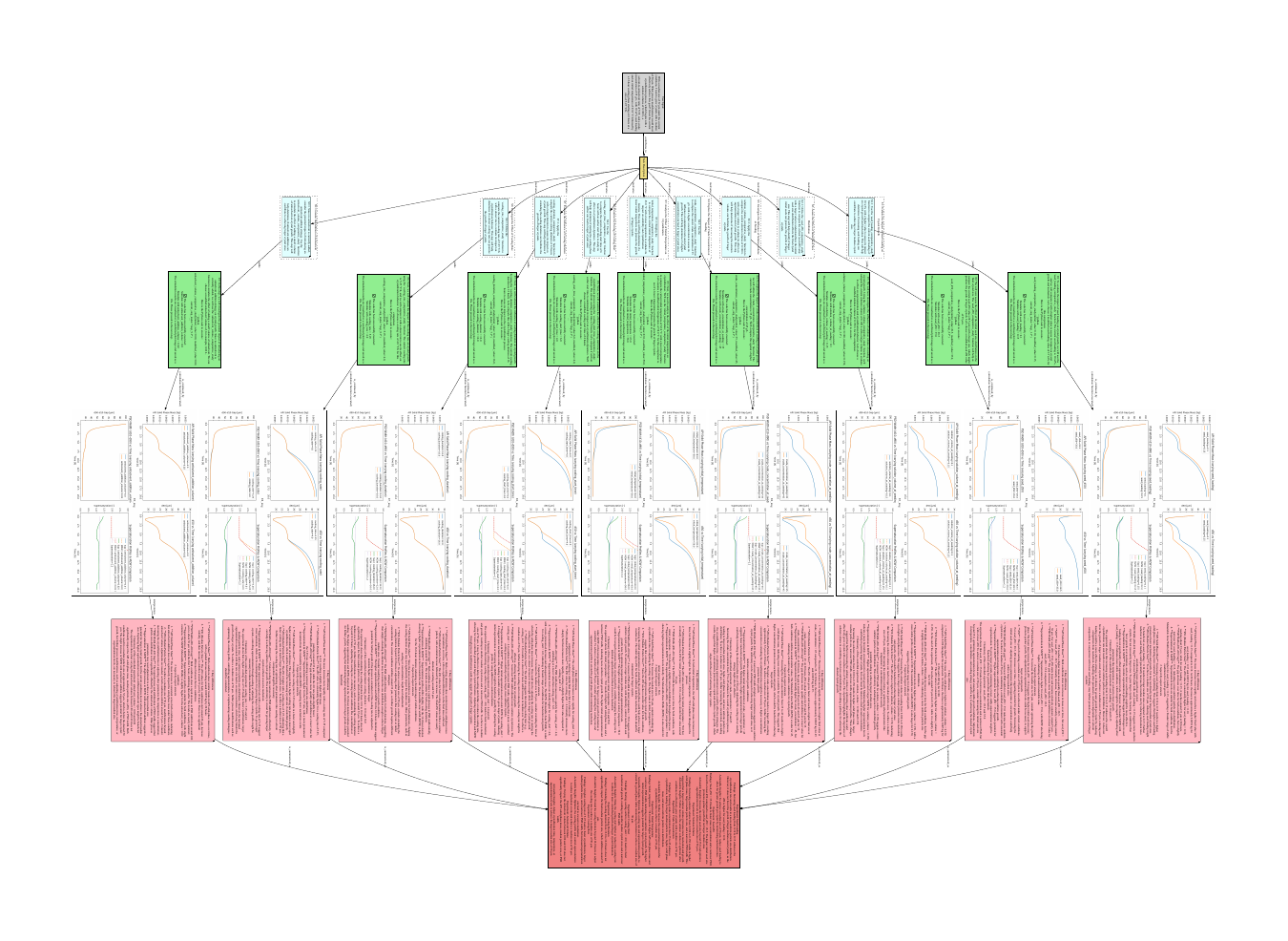}
}

\noindent\textbf{Task Sample 1 (Ablation: No Simulation integration)}

\noindent\includegraphics[width=\columnwidth]{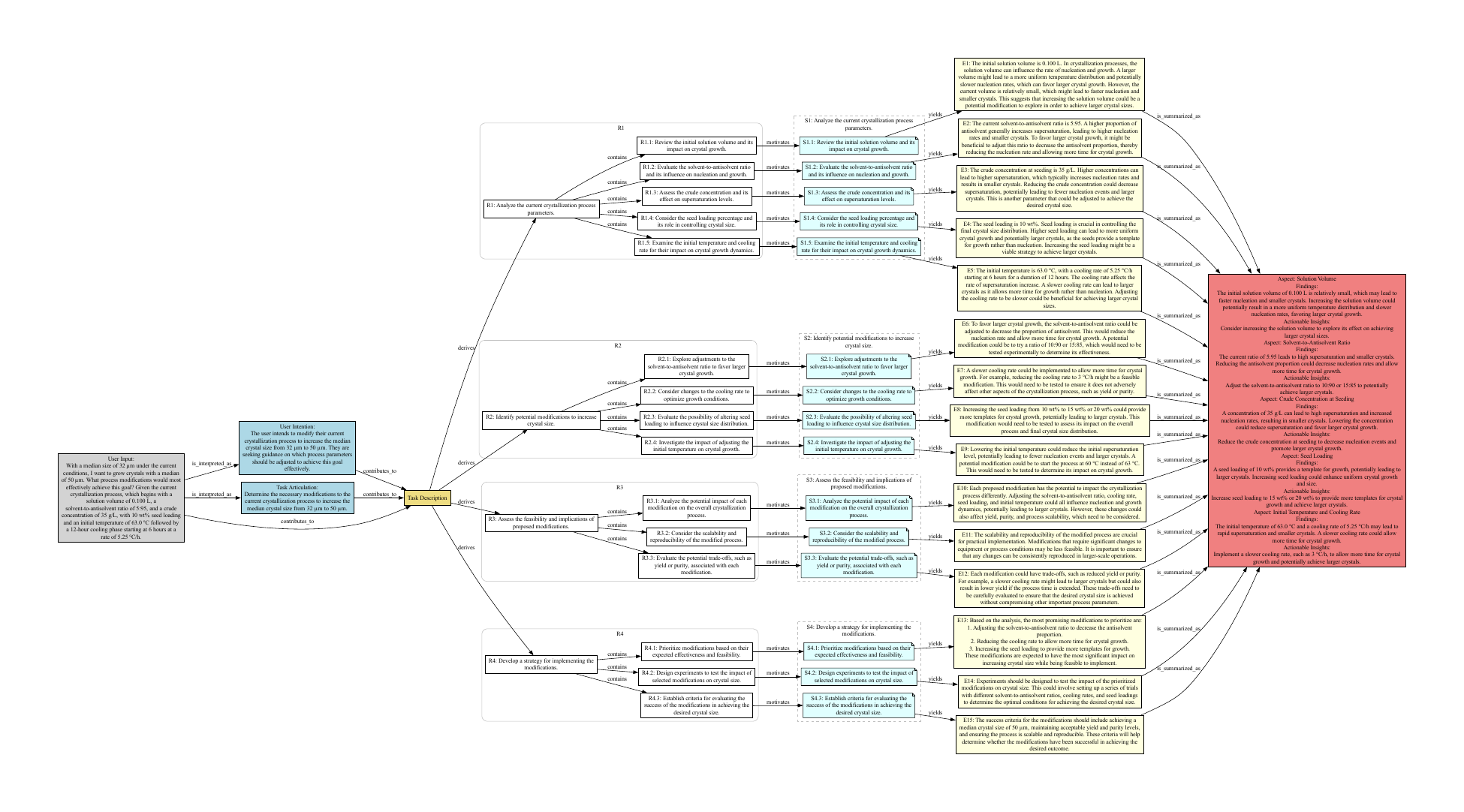}

\noindent\textbf{Task Sample 2 (Full System)}
\vspace{-1em}

\noindent\includegraphics[width=\columnwidth]{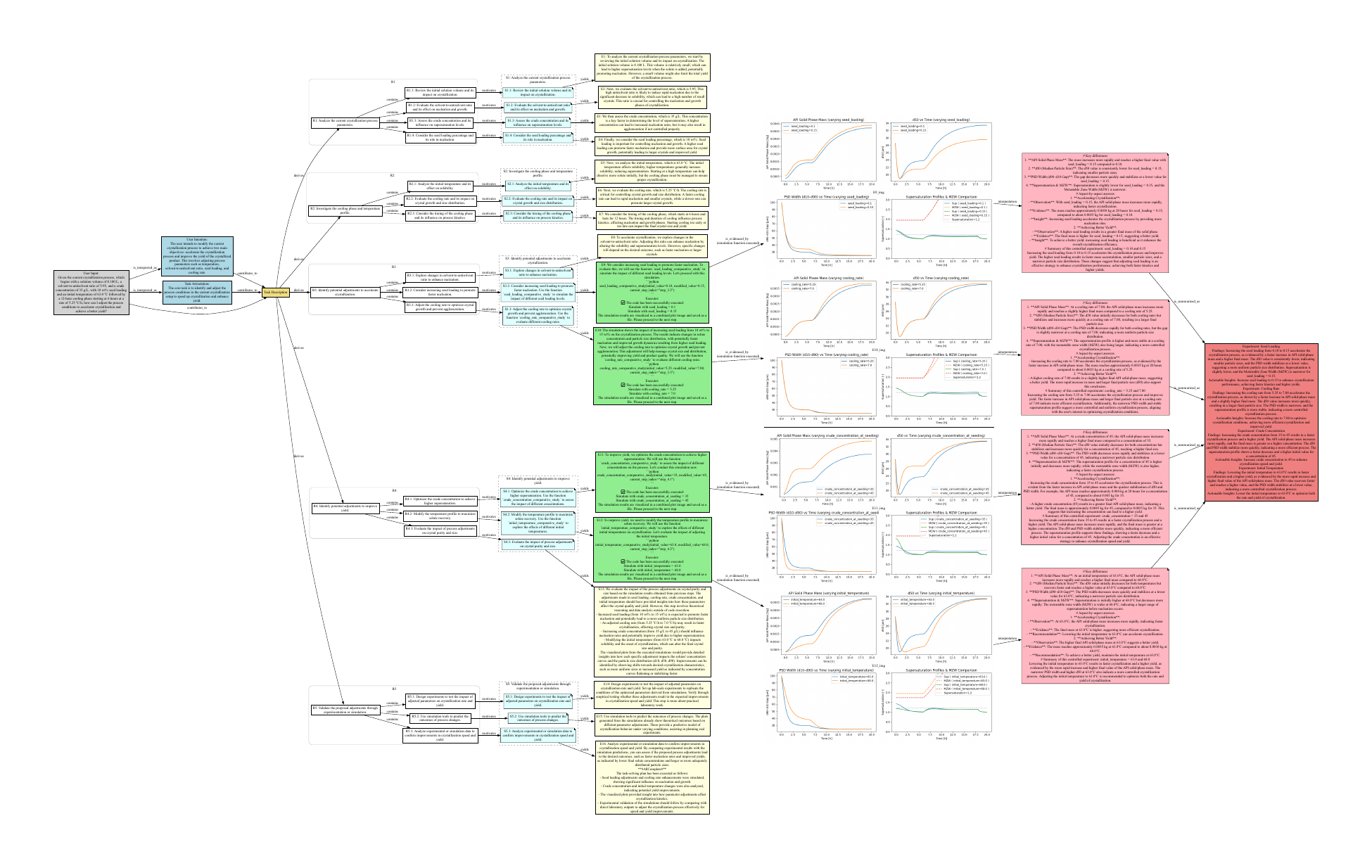}
\vspace{-2em}

\noindent\textbf{Task Sample 3 (Full System)}
\vspace{-1em}

\noindent\includegraphics[width=\columnwidth]{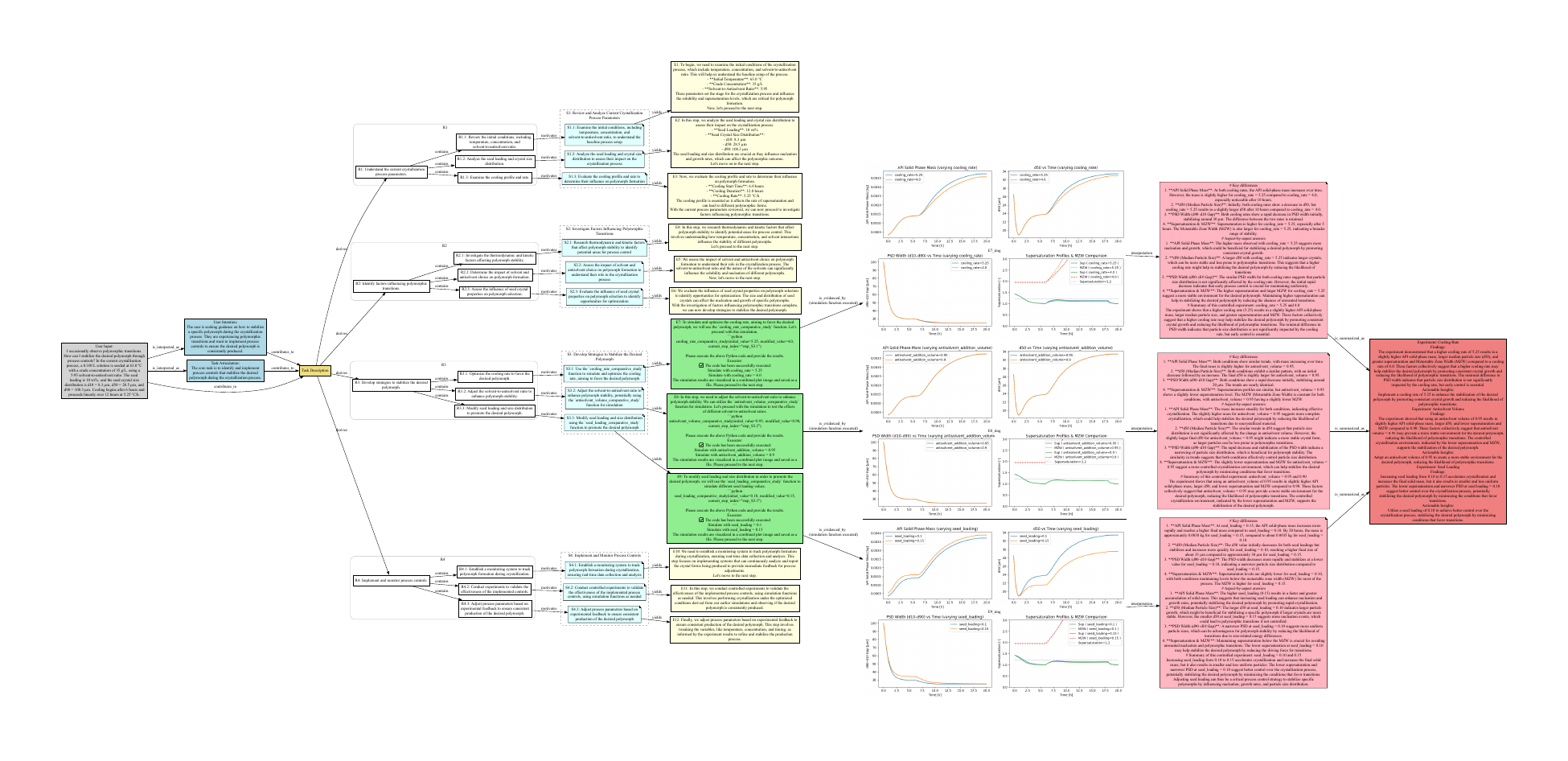}
\vspace{-2em}

\noindent\textbf{Task Sample 4 (Full System)}
\vspace{-1em}

\noindent\includegraphics[width=\columnwidth]{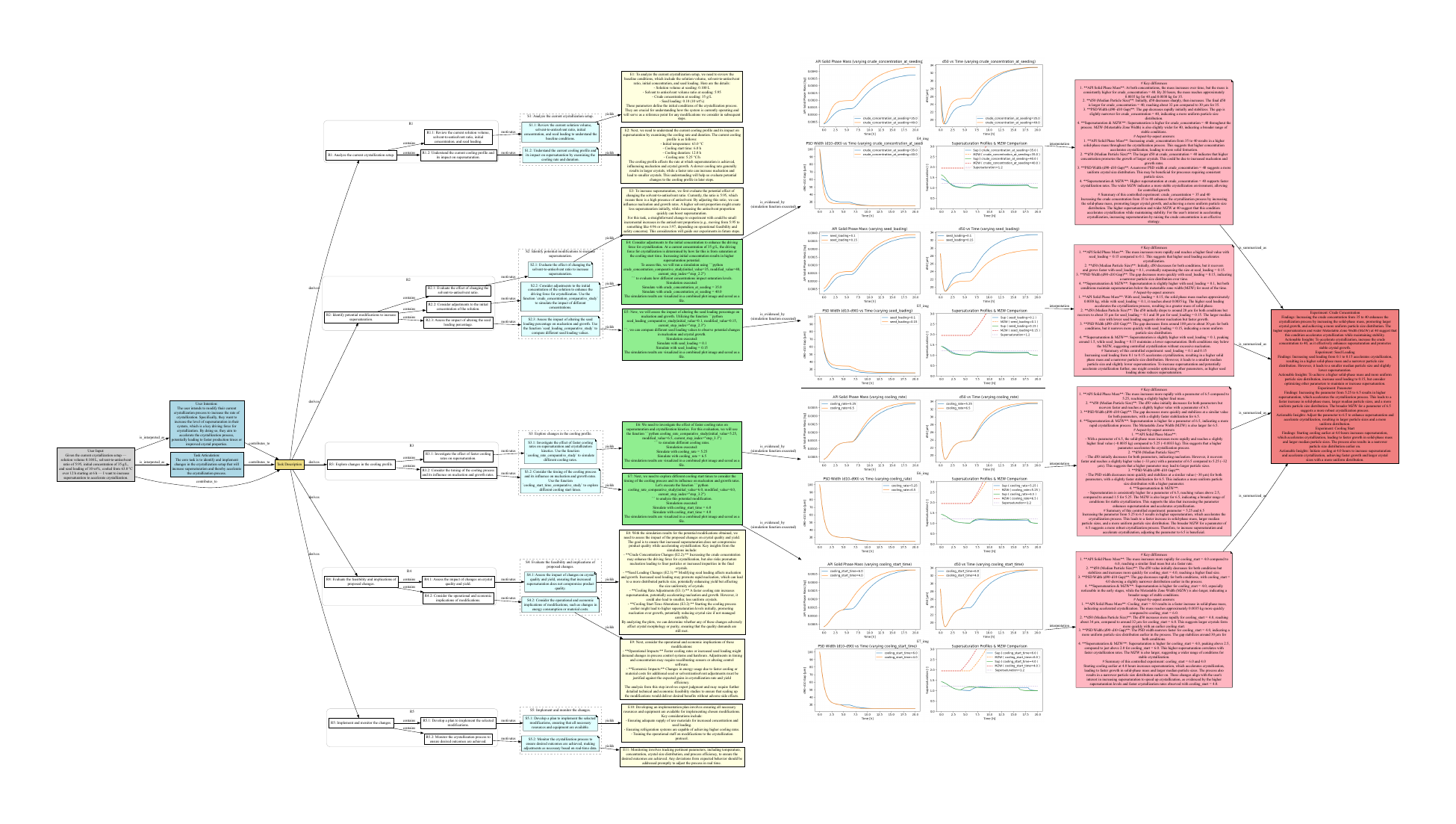}
\vspace{-2em}

\noindent\textbf{Task Sample 5 (Full System)}
\vspace{-1em}

\noindent\includegraphics[width=\columnwidth]{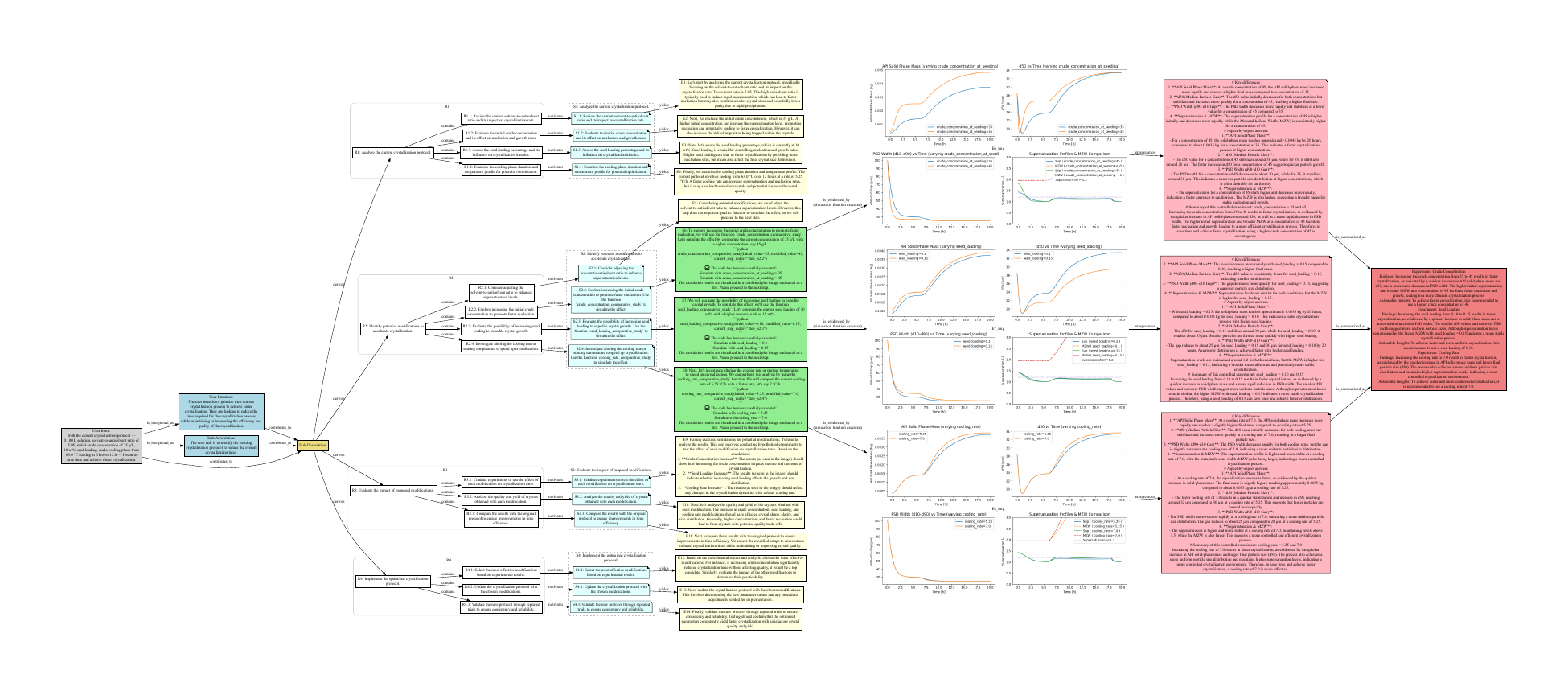}
\vspace{-2em}

\bibliographystyle{IEEEtran}
\bibliography{references}
\end{document}